\documentclass{article} 
\usepackage{iclr2025_conference,times}

\usepackage{amsmath,amsfonts,bm}

\def\eqref#1{equation~\ref{#1}}

\def\1{\bm{1}}

\DeclareMathAlphabet{\mathsfit}{\encodingdefault}{\sfdefault}{m}{sl}
\SetMathAlphabet{\mathsfit}{bold}{\encodingdefault}{\sfdefault}{bx}{n}

\usepackage{hyperref}
\usepackage{url}
\usepackage{graphicx}
\usepackage{booktabs}
\usepackage{tablefootnote} 

\usepackage{pgfplots}
\usepackage{tikz}
\usepackage{subcaption}
\usepackage{standalone}
\usepackage{wrapfig}

\usepackage{multirow}
\usepackage{pgfplotstable}
\usepackage{xcolor}
\usepackage{placeins}

\title{Training Domain Draft Models for Speculative Decoding: {\fontsize{14}{9}\selectfont Best Practices and Insights}}

\author{Fenglu Hong, Ravi Raju, Jonathan Lingjie Li, Bo Li, Urmish Thakker,\\
\textbf{Avinash Ravichandran, Swayambhoo Jain, Changran Hu}\\
SambaNova Systems, Inc. \\
Palo Alto, CA, USA \\
{\small \texttt{\{fenglu.hong,ravi.raju,jonathan.li,bo.li,urmish.thakker,}}\\
{\small \texttt{avinash.ravichandran,swayambhoo.jain,changran.hu\}@sambanovasystems.com}}
}

\makeatletter
\renewcommand{\footnotesize}{\scriptsize} 
\makeatother

\iclrfinalcopy 
\begin{document}

\maketitle
\begin{abstract}
Speculative decoding is an effective method for accelerating inference of large language models (LLMs) by employing a small draft model to predict the output of a target model. However, when adapting speculative decoding to \textbf{domain-specific} target models, the acceptance rate of the \textbf{generic} draft model drops significantly due to domain shift. In this work, we systematically investigate knowledge distillation techniques for \textbf{training domain draft models} to improve their speculation accuracy. We compare white-box and black-box distillation approaches and explore their effectiveness in various data accessibility scenarios, including historical user queries, curated domain data, and synthetically generated alignment data. Our experiments across \textit{Function Calling}, \textit{Biology}, and \textit{Chinese} domains show that offline distillation consistently outperforms online distillation by 11\% to 25\%, white-box distillation surpasses black-box distillation by 2\% to 10\%, and data scaling trends hold across domains. Additionally, we find that synthetic data can effectively align draft models and achieve 80\% to 93\% of the performance of training on historical user queries. These findings provide practical guidelines for training domain-specific draft models to improve speculative decoding efficiency.

\end{abstract}

\section{Introduction}

Large language models (LLMs) like GPT-4 \citep{openai2024gpt4technicalreport} and DeepSeek-R1 \citep{deepseekai2025deepseekr1incentivizingreasoningcapability} have demonstrated remarkable capabilities, but come with high computational costs and inference latency, limiting real-time applications. As a solution, speculative decoding accelerates model inference by using a smaller model (known as the draft model) to generate candidate outputs, which are then verified by the larger model (known as the target model) \citep{leviathan2023fastinferencetransformersspeculative, chen2023acceleratinglargelanguagemodel}. However, the performance of speculative decoding heavily depends on draft-target model alignment — misalignment increases verification failures, requiring the draft model to regenerate tokens. 
\begin{wraptable}{r}{0.44\textwidth} 
    \centering
    \renewcommand{\arraystretch}{1.1} 
    \begin{tabular}{l@{\hskip 1pt}c@{\hskip 1pt}c@{\hskip 1pt}c}

        \toprule
        & \multicolumn{1}{c}{Generic} & \multicolumn{1}{c}{Domain} & \multicolumn{1}{c}{Relative} \\
        & \multicolumn{1}{c}{Target} & \multicolumn{1}{c}{Target} & \multicolumn{1}{c}{Drop} \\
        \midrule
        Biology  & 60.7  & 37.5  & -38.2\% \\
        Chinese  & 49.6  & 35.7  & -28.0\% \\
        Coding   & 59.4  & 55.2  & -7.1\%  \\
        Math     & 78.3  & 75.4  & -3.7\%  \\
        \bottomrule
    \end{tabular}
    \caption{Avg. token acceptance rate (\%) for general target model and domain target model, using a general small model as draft model. Details in Appendix~\ref{sec:appendix AR drop}.}
    \label{tab:performance drop}
\end{wraptable}
Our experiments reveal that, when replacing a generic target model with a domain-specific fine-tuned model in speculative decoding, the speculation accuracy of the draft model in terms of the average token acceptance rate drops significantly in domain-specific queries, as shown in Table \ref{tab:performance drop}. This degradation underscores the need for domain-adapted draft models to maintain efficiency in speculative decoding.

To improve speculative decoding efficiency for domain target models, we explore various knowledge distillation techniques for training domain draft models \citep{agarwal2024onpolicydistillationlanguagemodels, hinton2015distillingknowledgeneuralnetwork, zhou2024distillspecimprovingspeculativedecoding} under various data constraints. We compare white-box distillation, which utilizes the target model parameters, and black-box distillation, which relies only on the target model outputs. We also examine training strategies under varying data availability conditions. Ideally, training with historical user queries ensures minimal domain shift and optimal performance. However, in many real-world scenarios, such data may not be available, particularly before a model has seen real user interaction \citep{liu2024onlinespeculativedecoding}. To address this, we evaluated alternatives, including training with curated domain queries and using synthetically generated data, such as Magpie \citep{xu2024magpiealignmentdatasynthesis}, from the target model. Our work  provides practical guidelines for training draft models under different constraints and thus improving speculative decoding efficiency in domain-specific applications.

In summary, our paper makes the following contributions:
\begin{itemize}
\item Comprehensive analysis of distillation methods: We compare white-box and black-box distillation for training draft models and evaluate their relative effectiveness across different domains.
\item Investigation of data accessibility constraints: We explore training strategies under three different data scenarios: (1) historical streaming/user query data, (2) collected domain-specific queries, and (3) synthetically generated data.
\item Empirical evaluation across multiple domains: We conduct experiments on three target domains (Function Calling, Biology, and Chinese) to assess the impact of different training methods on speculative decoding performance.
\item Guidance for practical draft model training: Our findings provide  insights on how to construct draft models under varying data constraints, offering a practical reference for improving inference efficiency in domain-specific LLM applications.
\end{itemize}

\begin{figure}[t]
\centering
\includegraphics[width=0.9\textwidth]{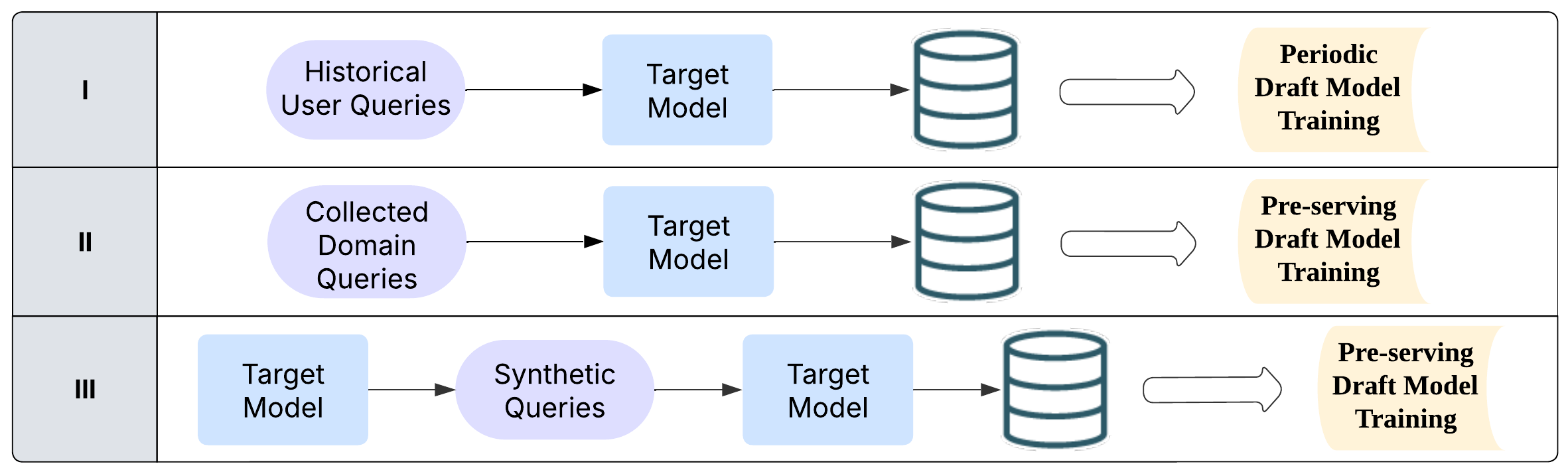} 
\caption{Three data accessibility scenarios for domain draft model training. Scenario I assumes access to historical user queries and train the draft model with distillation losses given target model's generations. Scenario II and III assume no access to use queries. We can use either collected domain queries (II) or synthetic queries generated by the target model (III) for training.}
\label{fig:Data scenarios}
\end{figure}

\section{Background \& Methods}

\subsection{Knowledge Distillation (KD) for speculative decoding} 
The effectiveness of speculative decoding depends on how well the output distribution from the draft model aligns with that of the target model. Knowledge distillation, which is a widely used framework for training a smaller student model to mimic the predictive distribution of a larger teacher model, is thus effective for enhancing speculative decoding \citep{zhou2024distillspecimprovingspeculativedecoding, liu2024onlinespeculativedecoding}. To develop more effective domain-specific draft models for speculative decoding, we use knowledge distillation techniques to improve the alignment between the target model ($M_p$) and draft model ($M_q$). We assume that the draft has learnable parameters $\theta$. We are also given a dataset of input sequences $X$. We generate \textbf{outputs from the target model} with greedy decoding, to form the training dataset $D = \{(x_k, p(x_k))\}^{|X|}_{k=1}$.


{\bf Supervised FT on target model's outputs} The draft model is finetuned to minimize the negative log-likelihood $L_{SFT}$ over the output sequences from the target model. In subsequent sections, this method is referred to as \textbf{SFT}.

\begin{equation}
L_{SFT}(\theta) := \mathbb{E}_{(x,y) \sim (X,Y)} \left[ -\log q_{\theta} (y | x) \right]
\end{equation}

{\bf Supervised KD on target model's outputs} This is a white-box distillation technique where the draft model is trained to mimic the token-level probability distributions of the target model. Specifically, the draft model is trained with 
\begin{equation}
    L_{KD}(\theta) := \mathbb{E}_{(x,y) \sim (X,Y)} \left[ \mathcal{D} \left( p \| q_{\theta} \right) (y | x) \right],
\end{equation}
where we use the forward Kullback Leibler divergence (KL) and reverse KL (RKL) for $\mathcal{D}$.

{\bf Online vs. Offline Distillation}
For the white-box distillation, we further investigate two learning paradigms: online distillation and offline distillation, as proposed in \citep{liu2024onlinespeculativedecoding}. In offline distillation setting, the draft model has unrestricted access to the static dataset $D$. 
Offline distillation does not allow real-time adaptation to shifts in data distribution. In contrast, online distillation refines the draft model dynamically during the speculative decoding inference process. Specifically, the draft model proposes tokens during inference, which are verified by the target model. The target logits and draft logits for the incorrect predictions are stored in a buffer, and the draft model is updated every time when the buffer exceeds a threshold.

\subsection{Magpie alignment data synthesis}
\textbf{Magpie} is a synthetic data generation technique used to generate training data for model alignment \citep{xu2024magpiealignmentdatasynthesis}. By providing the (aligned) target model with a pre-query template (and potentially a system prompt), the model generates both sample queries and corresponding completions. We re-purpose Magpie for creating draft model training data for speculative decoding. This approach offers two key advantages: (1) it is data-free, requiring no pre-existing datasets, and (2) it eliminates biases associated with selecting training data.

\section{Experiments}

\subsection{Experimental Setup}
We adopt the setup presented in {\it Online Speculative Decoding} paper \citep{liu2024onlinespeculativedecoding} for white-box distillation and evaluate our methods primarily on the LLaMA series \citep{grattafiori2024llama3herdmodels}. Specifically, we use the LLaMA 3.2 1B Instruct model\footnote{\url{https://huggingface.co/meta-llama/Llama-3.2-1B-Instruct}} as the draft model and, unless otherwise specified, conduct all experiments with LLaMA 3.1 8B-sized target models. 
To assess the effectiveness of distillation approaches, we focus on adapting the draft model to niche domains; we select 1) Function Calling, 2) Biology, and 3) Chinese to ensure sufficient coverage. For each domain, we select a domain-specific target model and use open-source domain datasets from Hugging face, setting aside 1000 prompts for the test set. To evaluate performance, we measure the \textit{average token acceptance rate} of the draft model on the test set with proposal length $k=9$. More details on target models, datasets, and training configurations are discussed in Appendix~\ref{sec:experiment setup}.

\subsection{Data Accessibility Scenarios}

We study three different data accessibility scenarios and corresponding data collection techniques for training domain-specific draft models, as depicted in Figure \ref{fig:Data scenarios}.

{\bf I. Using historical user query data} This scenario assumes minimal domain shift between training and deployment. To simulate such a setting, we select a domain-specific dataset $D$ and apply a train-test split for training the draft model and evaluating it under speculative decoding setting.

{\bf II. Using collected domain-specific queries} To simulate scenario where the user query data is unavailable but related domain queries exist, we train the draft model on dataset $D'$ and evaluate it on a separate dataset $D$. Both datasets belong to the same domain, but their queries may exhibit minor domain shifts, providing insight into the robustness of draft model training under distributional variations.

{\bf III. Using synthetically generated queries} Collecting domain queries often requires significant human effort. However, when user queries and curated domain queries are unavailable, we leverage the target model to generate synthetic instructions and corresponding completions. Specifically, we adopt the synthetic data generation method proposed by \citep{xu2024magpiealignmentdatasynthesis}, which eliminates the need for prompt engineering or manually curated seed instructions.

\section{Results and Discussion}
\begin{figure}[h]
\centering
\includegraphics[width=0.86\textwidth]{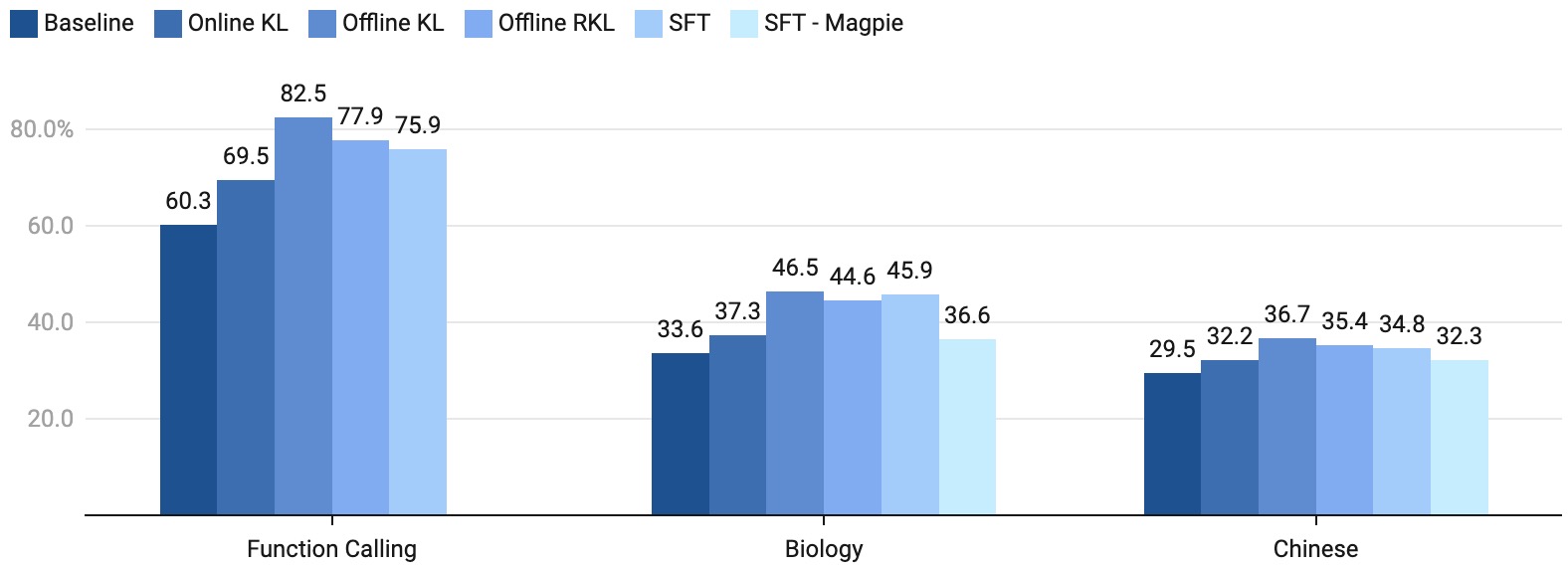} 
\caption{Average acceptance rates for different methods. All methods (except SFT - Magpie) train with in-domain data where a domain-specific dataset is split into training and test sets, mimicking real user queries (Scenario I). SFT - Magpie method trains with Magpie synthetic data (Scenario III). More details in Appendix \ref{sec:appendix main results}.}
\label{fig:main results}
\end{figure}

\vspace{-0.1cm}
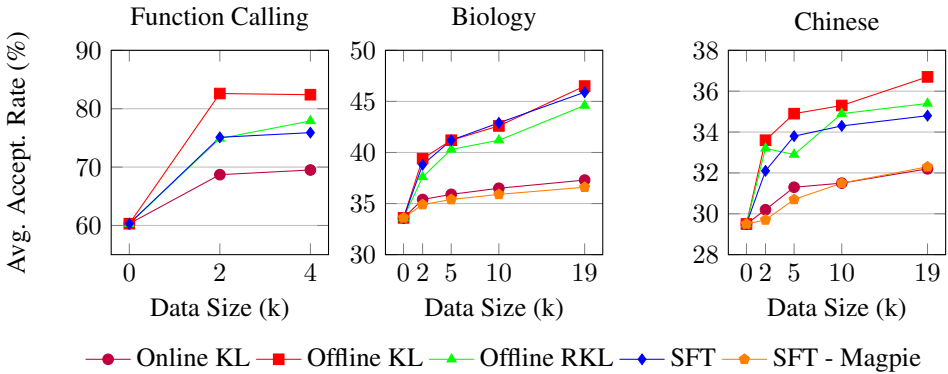
\begin{figure}[ht]
    \centering
    \centering
\begin{subfigure}{0.32\textwidth}
    \begin{tikzpicture}
        \begin{axis}[
            width=\textwidth,
            height=4.3cm,
            title={Function Calling},
            title style={yshift=-2pt},
            xlabel style={yshift=2pt},
            xlabel={Data Size (k)},
            ylabel={Avg. Accept. Rate (\%)},
            xtick={0,2,4},
            ymin=55, ymax=90, ymajorgrids
        ]
            \addplot[color=purple, mark=*] plot coordinates {(0,60.3) (2,68.7) (4,69.5)};
            \addplot[color=red, mark=square*] plot coordinates {(0,60.3) (2,82.6) (4,82.4)};
            \addplot[color=green, mark=triangle*] plot coordinates {(0,60.3) (2,74.9) (4,77.9)};
            \addplot[color=blue, mark=diamond*] plot coordinates {(0,60.3) (2,75.1) (4,75.9)};
        \end{axis}
    \end{tikzpicture}
\end{subfigure}
\begin{subfigure}{0.32\textwidth}
    \begin{tikzpicture}
        \begin{axis}[
            width=\textwidth,
            height=4.3cm,
            title={Biology},
            title style={yshift=-2pt},
            xlabel style={yshift=2pt},
            xlabel={Data Size (k)},
            ylabel={ },
            xtick={0,2,5,10,19},
            ymin=30, ymax=50, ymajorgrids
        ]
            \addplot[color=purple, mark=*] plot coordinates {(0,33.6) (2,35.4) (5,35.9) (10,36.5) (19,37.3)};
            \addplot[color=red, mark=square*] plot coordinates {(0,33.6) (2,39.4) (5,41.2) (10,42.6) (19,46.5)};
            \addplot[color=green, mark=triangle*] plot coordinates {(0,33.6) (2,37.6) (5,40.3) (10,41.2) (19,44.6)};
            \addplot[color=blue, mark=diamond*] plot coordinates {(0,33.6) (2,38.8) (5,41.2) (10,42.9) (19,45.9)};
            \addplot[color=orange, mark=pentagon*] plot coordinates {(0,33.6) (2,34.9) (5,35.4) (10,35.9) (19,36.6)};
        \end{axis}
    \end{tikzpicture}
\end{subfigure}
\begin{subfigure}{0.32\textwidth}
    \begin{tikzpicture}
        \begin{axis}[
            width=\textwidth,
            height=4.3cm,
            title={Chinese},
            title style={yshift=-2pt},
            xlabel style={yshift=2pt},
            xlabel={Data Size (k)},
            ylabel={ },
            xtick={0,2,5,10,19},
            ymin=28, ymax=38, ymajorgrids
        ]
            \addplot[color=purple, mark=*] plot coordinates {(0,29.5) (2,30.2) (5,31.3) (10,31.5) (19,32.2)};
            \addplot[color=red, mark=square*] plot coordinates {(0,29.5) (2,33.6) (5,34.9) (10,35.3) (19,36.7)};
            \addplot[color=green, mark=triangle*] plot coordinates {(0,29.5) (2,33.2) (5,32.9) (10,34.9) (19,35.4)};
            \addplot[color=blue, mark=diamond*] plot coordinates {(0,29.5) (2,32.1) (5,33.8) (10,34.3) (19,34.8)};
            \addplot[color=orange, mark=pentagon*] plot coordinates {(0,29.5) (2,29.7) (5,30.7) (10,31.5) (19,32.3)};
        \end{axis}
    \end{tikzpicture}
\end{subfigure}

\vspace{-6.2cm}
\begin{tikzpicture}
    \begin{axis}[
        hide axis,
        xmin=0, xmax=1, ymin=0, ymax=1,
        legend columns=5,
        legend style={draw=none, at={(0.5,-0.1)}, anchor=north}
    ]
        \addlegendimage{color=purple, mark=*}
        \addlegendentry{Online KL}
        \addlegendimage{color=red, mark=square*}
        \addlegendentry{Offline KL}
        \addlegendimage{color=green, mark=triangle*}
        \addlegendentry{Offline RKL}
        \addlegendimage{color=blue, mark=diamond*}
        \addlegendentry{SFT}
        \addlegendimage{color=orange, mark=pentagon*}
        \addlegendentry{SFT - Magpie}
    \end{axis}
    \vspace{-0.2cm}
\end{tikzpicture}
    \caption{Performance scales with dataset size. Besides, as training data increases, the offline KL approach gains an increasing advantage over online KL in Biology and Chinese domains.}
    \label{fig:data_scaling}
\end{figure}

In this section, we present our experimental findings and analyze the impact of different distillation methods on speculative decoding performance across various domains.

{\bf Offline vs. Online Distillation} 
Our experiments show that offline distillation consistently outperforms online distillation across all three domains (see Figure~\ref{fig:main results}), and this trend holds across different dataset sizes (see Figure~\ref{fig:data_scaling}). A key insight into this trend is that offline distillation leverages supervision from all completion tokens, providing richer learning signals for the draft model. Furthermore, as training data increases, the offline approach gains a growing advantage over online approach in Biology and Chinese. In Biology, offline surpasses online by 11.4\% to 24.7\% in acceptance rate as data expands from 2k to 19k; in Chinese, the gap widens from 11.2\% to 14.1\%. 
We also find out that for training with in-domain user queries data (Data Scenario I), offline distillation can benefit from higher learning rate while online distillation requires a lower learning rate, as detailed in Appendix~\ref{sec:optimal lrs for online vs. offline}. However, for Data Scenario II and III where the training data exhibits domain shifts from evaluation data, offline distillation would also prefer a lower learning rate (see Table ~\ref{tab:magpie_offline} and~\ref{tab:apigen}).

\begin{table}[t]
    \centering
    \begin{tabular}{l l c}
        \toprule
        \textbf{Target Model} & \textbf{Draft Model} & \textbf{Biology} \\
        \midrule
        Llama3-OpenBioLLM-70B\tablefootnote{\url{https://huggingface.co/aaditya/Llama3-OpenBioLLM-70B}} & Llama3.2-1B\tablefootnote{\url{https://huggingface.co/meta-llama/Llama-3.2-1B-Instruct}} & 33.0\% \\
             & \ + SFT (Data Scenario I) & +8.4\% \\
        \bottomrule
    \end{tabular}
    \caption{Accept. rate(\%) of 1B \textbf{SFT}-trained draft model for decoding 70B target model in Biology domain. Training data size is 4k.}
    \label{tab:70b_bio}
\end{table}

\begin{table}[t]
    \centering
    \begin{tabular}{l l c c c c}
        \toprule
        \textbf{Target Model} & \textbf{Draft Model} & \textbf{Math} & \textbf{Coding} & \textbf{General} & \textbf{Avg.} \\
        \midrule
        DS-R1-Qwen-32B\tablefootnote{\url{https://huggingface.co/deepseek-ai/DeepSeek-R1-Distill-Qwen-32B & qwen2.5-0.5B-Instruct}} & Qwen2.5-0.5B\tablefootnote{\url{https://huggingface.co/Qwen/Qwen2.5-0.5B-Instruct}} & 34.7\% & 35.6\% & 34.6\% & 35.0\% \\
                    & \ + SFT (Data Scenario II) & +5.0\% & +5.0\% & +5.0\% & +5.0\% \\
        \midrule
        DS-R1-Llama-70B\tablefootnote{\url{https://huggingface.co/deepseek-ai/DeepSeek-R1-Distill-Llama-70B}} & Llama3.2-1B\tablefootnote{\url{https://huggingface.co/meta-llama/Llama-3.2-1B-Instruct}} & 41.6\% & 32.3\% & 26.9\% & 33.6\% \\
                     & \ + SFT (Data Scenario II) & +10.0\% & +17.7\% & +6.0\% & +11.2\% \\
        \bottomrule
    \end{tabular}
    \caption{Accept. rate(\%) of \textbf{SFT}-trained draft models for decoding 70B and 32B target models in Reasoning domain. Training data size is 200k.}
    \label{tab:reasoning}
\end{table}

\begin{table}[t]
    \vspace{-0.2cm}
    \centering
    \begin{tabular}{l c c c}
        \toprule
        & \textbf{Data Size} & \textbf{LR=1e-6} & \textbf{LR=2e-5} \\
        \midrule
        Offline KL on Chinese train set (Data Scenario I) & 19k  & 35.6 & \textbf{36.7} \\
        Offline KL on Magpie syn. data (Data Scenario III) & 19k & \textbf{33.2} & 31.7 \\
        \bottomrule
    \end{tabular}
    \vspace{-0.1cm}
    \caption{Avg. accept. rate (\%) on Chinese test set. The baseline accept. rate is 29.5\%. Offline KL with Magpie synthetic data achieves 90.4\% of the performance of training with in-domain data.}
    \label{tab:magpie_offline}
\end{table}

{\bf Data Scaling Law}
As shown in Figure~\ref{fig:data_scaling}, we observe that the data scaling law generally holds across all domains and training methods, with larger datasets yielding better draft model alignment. One exception occurs in Function Calling domain with offline distillation using forward KL loss, where 2000 training samples already reach optimal performance. This is likely due to the highly structured nature of function-calling outputs, which require less training data to achieve alignment.

{\bf White-Box vs. Black-Box Distillation}
White-box offline distillation with forward KL loss generally outperforms black-box distillation (SFT) (see Figure \ref{fig:main results} and \ref{fig:data_scaling}), indicating that leveraging the target model’s logits provides a stronger training signal than relying solely on final output tokens.
In our investigated domains, the former achieves 1.3\% $\sim$ 9.9\% more acceptance rate than SFT.

\textbf{Effectiveness of Synthetic Data (Magpie)}
We examine the effectiveness of Magpie synthetic data for aligning draft models in Biology and Chinese domains. While not as effective as training on in-domain data, synthetic data still yields meaningful improvements (see Figure \ref{fig:main results}), making it a viable approach for training an initial draft model before target model deployment. Notably, the performance gap between user query finetuning and synthetic data finetuning is larger in Biology than in Chinese. This can be attributed to the fact that the biomedical target model primarily generating diagnostic and medical-related completions, which exhibit a greater domain shift from real user queries that span biological topics. We also find that offline training with Magpie data benefits from a smaller learning rate than in-domain user query data and achieves over 90\% of the acceptance rate observed with in-domain training (see Table~\ref{tab:magpie_offline}). This is likely because Magpie data exhibits some domain shift from the evaluation data, and thus a larger LR leads to overfitting.

\textbf{Scaling to Larger Target Model} In order to assess the effectiveness of training draft models for larger target models, we extend our experiments to models exceeding 8B parameters in the domains of Biology and Reasoning. As presented in Table~\ref{tab:70b_bio}, within the Biology domain, SFT with 4,000 training samples improves the acceptance rate of the 1B draft model by 25.4\% for the 70B target model (Llama3-OpenBioLLM-70B). 

Additionally, we evaluate our method for two reasoning models distilled from DeepSeek-R1 \citep{deepseekai2025deepseekr1incentivizingreasoningcapability}, which incorporate extended reasoning processes in their outputs and demonstrate strong performance in reasoning-intensive tasks. To conduct our experiments within the framework of Data Scenario II, we use a multi-domain instruction dataset for training and reasoning-intensive domain datasets for evaluation. Specifically, we sample 200,000 prompts from a combination of the training dataset introduced in \cite{jain2024compositionexpertsmodularcompound} and the Magpie-500K dataset\footnote{\url{https://huggingface.co/datasets/Magpie-Align/Magpie-Llama-3.1-Pro-500K-Filtered}}, covering math, coding, and additional domains. For evaluation, we construct domain-specific datasets for \textit{math}, \textit{coding}, and \textit{general} reasoning by sampling 500 prompts from AIME\footnote{\url{https://huggingface.co/datasets/di-zhang-fdu/AIME\_1983\_2024}}, BigCodeBench\footnote{\url{https://huggingface.co/datasets/bigcode/bigcodebench}}, and AlpacaEval \citep{dubois2024alpacafarmsimulationframeworkmethods}, respectively. As summarized in Table~\ref{tab:reasoning}, the SFT-trained Qwen 2.5 0.5B Instruct model achieves a 14.3\% improvement in acceptance rate when serving as a draft model for the 32B reasoning model. Similarly, the SFT-trained LLaMA 3.2 1B Instruct model exhibits a 33.3\% improvement when used as a draft model for the 70B reasoning model. These results indicate that the effectiveness of SFT extends to larger target models, and training on collected domain-relevant data can significantly enhance the alignment of draft models.

\begin{table}[t]
    \centering
    \begin{tabular}{l c c c}
        \toprule
        & \textbf{Data Size} & \textbf{LR=1e-6} & \textbf{LR=2e-5} \\
        \midrule
        Offline KL on Hermes-FC\tablefootnote{\url{https://huggingface.co/datasets/NousResearch/hermes-function-calling-v1}} (Data Scenario I) & 2k  & 76.4 & \textbf{82.6} \\
        Offline KL on APIGen-FC\tablefootnote{\url{https://huggingface.co/datasets/argilla/apigen-function-calling}} (Data Scenario II) & 20k & \textbf{72.1} & 63.6 \\
        \bottomrule
    \end{tabular}
    \vspace{-0.1cm}
    \caption{Avg. accept. rate(\%) on the test split of Hermes-FC dataset. The baseline accept. rate is 60.3\%. Training on related domain data reaches 87.3\% of the performance of using in-domain data.}
    \label{tab:apigen}
    \vspace{-0.1cm}
\end{table}

\textbf{Training on Related Domain Data}
For Data Scenario II, when user queries are unavailable, training on related domain data also effectively improves draft model alignment. In Function Calling domain, training on APIGen data\footnote{\url{https://huggingface.co/datasets/argilla/apigen-function-calling}} enhances the draft model's performance on Hermes-FC\footnote{\url{https://huggingface.co/datasets/NousResearch/hermes-function-calling-v1}} (Table~\ref{tab:apigen}), which serves as a proxy for real user queries. However, domain shift between collected data and user queries necessitates significantly more training samples to achieve comparable results and more careful learning rate selection. Beyond the Function Calling domain, as discussed in the previous section, SFT on collected domain datasets also greatly improves the performance of the draft models in math, coding, and general reasoning tasks (see Table~\ref{tab:reasoning}).


\section{Conclusion}
This work investigates best practices for training domain-specific draft models to improve speculative decoding efficiency when paired with specialized target models. Our experiments show that offline knowledge distillation outperforms online learning by 11\% to 25\%, with forward KL loss providing the optimal result. We also demonstrate that white-box distillation, which utilizes target model logits, exceeds the black-box approach by 2\% to 10\%. Additionally, we explore data accessibility scenarios and find that synthetic alignment data can achieve 80\% to 93\% of the performance of training on in-domain data. These insights provide actionable guidelines for the construction of effective draft models under different constraints, ultimately enhancing speculative decoding for domain-specific applications. 
\newpage

\bibliography{iclr2025_conference}
\bibliographystyle{iclr2025_conference}

\newpage

\appendix

\appendix

\section{Related Work}
\subsection{Speculative Decoding for LLMs}
Speculative decoding (SD) is a means to accelerate LLM decoding by leveraging a small draft model to predict the outputs from a large target model. These candidate output tokens are then validated by a larger target model \citep{leviathan2023fastinferencetransformersspeculative, chen2023acceleratinglargelanguagemodel}. This decoupling of candidate generation from verification permits a reduction in the number of target model invocations, thereby decreasing overall inference time.

Most prior work in speculative decoding focuses on accelerating general-purpose LLMs by pairing them with small, general-purpose draft models—even when evaluations are performed on domain-specific tasks. In contrast, our study systematically investigates how to employ knowledge distillation to adapt a general-purpose draft model for use with a domain-specific target model. This work provides practical guidelines and best practices for aligning the draft model to the specialized output distribution, thereby reducing inference latency while preserving domain-specific performance.

\subsection{Knowledge Distillation for LLMs}
Knowledge distillation (KD) has long been recognized as an effective strategy for compressing large models into smaller, more efficient ones without a substantial loss in performance \citep{hinton2015distillingknowledgeneuralnetwork}. In the context of LLMs, KD enables a smaller “student” model to learn from a larger “teacher” model, thereby significantly reducing inference cost while striving to maintain high-quality output generation.

Early applications of KD in LLMs have primarily focused on black-box distillation \citep{alpaca}, where the student model is trained solely on the teacher’s outputs, often accessed via APIs. While practical when teacher internals are unavailable, this approach inherently limits the granularity of supervision. Recent advances have introduced white-box KD \citep{zhou2024distillspecimprovingspeculativedecoding, agarwal2024onpolicydistillationlanguagemodels, gu2024minillmknowledgedistillationlarge, wen-etal-2023-f}, which leverages internal representations such as logits and attention maps from the teacher to provide richer supervisory signals and, consequently, improved student performance.

\section{Experiment Details}
\label{sec:experiment setup}
In this section, we provide detailed information on our experiment setup. 

\subsection{Domain datasets and target models} 
\label{sec:domain-datasets-models}
This section includes information about the target model selection and dataset selection for training and evaluation. 

We select one Llama 3.1 8B-sized target model for each domain that we study. Specifically, we use Hermes-3-Llama-3.1-8B model for Function Calling\footnote{\url{https://huggingface.co/NousResearch/Hermes-3-Llama-3.1-8B}}, the Llama-3.1-8B-UltraMedical model for the Biology domain\footnote{\url{https://huggingface.co/TsinghuaC3I/Llama-3.1-8B-UltraMedical}}, and the Llama3.1-8B-Chinese-Chat for the Chinese domain\footnote{\url{https://huggingface.co/shenzhi-wang/Llama3.1-8B-Chinese-Chat}}. 

For the domain-specific datasets that are used to mimic real user queries (Scenario I), we make the following selection:
\begin{itemize}
    \item Function Calling: We use {\it NousResearch/hermes-function-calling-v1}\footnote{\url{https://huggingface.co/datasets/NousResearch/hermes-function-calling-v1}} dataset (Hermes-FC) and exclude Subset "function\_calling", which consists of multi-turn conversation function calls.
    \item Biology: We use {\it camel-ai/biology}\footnote{\url{https://huggingface.co/datasets/camel-ai/biology}} dataset.
    \item Chinese: We use {\it allenai/WildChat-1M}\footnote{\url{https://huggingface.co/datasets/allenai/WildChat-1M}} dataset, and filter for turn=1 and language = Chinese.
    
\end{itemize}
We reserve 1000 prompts in each dataset for the test set, and sample subsets of the remaining prompts as domain training data.

For Function Calling domain, we investigate data accessibility scenario II. We use {\it argilla/apigen-function-calling}\footnote{\url{https://huggingface.co/datasets/argilla/apigen-function-calling}} dataset (APIGen-FC) as domain dataset $D'$, which exhibits some domain shift from Hermes-FC dataset but both belong to Function Calling domain.

\subsection{Training Hyperparameters}
\begin{table}[h]
    \centering
    \begin{tabular}{l c c c}
        \toprule
        & \textbf{Batch Size} & \textbf{\# Epochs} & \textbf{Learning Rate} \\
        \midrule
        Online  & 8 & 1 & 1e-6 \\
        Offline & 8 & 3 & 2e-5 \\
        SFT     & 8 & 3 & 2e-5 \\
        \bottomrule
    \end{tabular}
    \caption{Training hyperparameters for different methods, if not otherwise specified for ablations.}
    \label{tab:training_configs}
\end{table}

\section{Additional Details on Tables \& Figures}
This section provides more details on some tables and figures in the main text.

\subsection{Table ~\ref{tab:performance drop}}
\label{sec:appendix AR drop}
For each line in Table \ref{tab:performance drop}, we use a small generic model as the draft model and compare its acceptance rates when decoding a general target model and a domain target model of the same size. The choice of models and evaluation domain datasets are specified in \ref{tab:AR_drop_models_datasets}. We randomly select 1000 prompts from each dataset as the evaluation set.
\begin{table}[h]
    \centering
    \renewcommand{\arraystretch}{1.2}
{\fontsize{8.2}{9}\selectfont
    \begin{tabular}{l c c c c}
        \toprule
        & \textbf{Generic Draft Model} & \textbf{Generic Target Model} & \textbf{Domain Target Model} & \textbf{Domain Data} \\
        \midrule
        \textbf{Biology} & Llama-3.1-8B\tablefootnote{\url{https://huggingface.co/meta-llama/Meta-Llama-3.1-8B-Instruct}} & Llama-3.1-70B\tablefootnote{\url{https://huggingface.co/meta-llama/Meta-Llama-3.1-70B-Instruct}} & OpenBioLLM-70B\tablefootnote{\url{https://huggingface.co/aaditya/Llama3-OpenBioLLM-70B}} & CAMEL-Bio\tablefootnote{\url{https://huggingface.co/datasets/camel-ai/biology}} \\
        \textbf{Chinese} & Llama-3.1-8B\tablefootnote{\url{https://huggingface.co/meta-llama/Meta-Llama-3.1-8B-Instruct}} & Llama-3.1-70B\tablefootnote{\url{https://huggingface.co/meta-llama/Meta-Llama-3.1-70B-Instruct}} & Llama3.1-70B-ZH\tablefootnote{\url{https://huggingface.co/shenzhi-wang/Llama3.1-70B-Chinese-Chat}} & WildChat\tablefootnote{\url{https://huggingface.co/datasets/allenai/WildChat}} Chinese split \\
        \textbf{Coding} & Qwen2.5-1.5B\tablefootnote{\url{https://huggingface.co/Qwen/Qwen2.5-1.5B-Instruct}} & Qwen2.5-7B\tablefootnote{\url{https://huggingface.co/Qwen/Qwen2.5-7B-Instruct}} & Qwen2.5-Coder-7B\tablefootnote{\url{https://huggingface.co/Qwen/Qwen2.5-Coder-7B-Instruct}} & BigCodeBench\tablefootnote{\url{https://huggingface.co/datasets/bigcode/bigcodebench}} \\
        \textbf{Math} & Qwen2.5-7B\tablefootnote{\url{https://huggingface.co/Qwen/Qwen2.5-7B-Instruct}} & Qwen2.5-72B\tablefootnote{\url{https://huggingface.co/Qwen/Qwen2.5-72B-Instruct}} & Qwen2.5-Math-72B\tablefootnote{\url{https://huggingface.co/Qwen/Qwen2.5-Math-72B-Instruct}} & Hendrycks-Math\tablefootnote{\url{https://huggingface.co/datasets/EleutherAI/hendrycks_math}} \\
        \bottomrule
    \end{tabular}
}
    \caption{Models and datasets selection.}
    \label{tab:AR_drop_models_datasets}
\end{table}

\subsection{Figure ~\ref{fig:main results}}
\label{sec:appendix main results}
Figure \ref{fig:main results} comprehensively compare the results from different training methods. For Biology and Chinese, the training sample size is 19k; for Function Calling, the training sample size is 4k. Online distillation uses LR=1e-6; offline distillation and SFT use LR=2e-5. For offline KL and SFT, evaluation is done after three-epoch training. 

\section{Ablations}

\subsection{Optimal LRs for Online vs. Offline Distillation}
\label{sec:optimal lrs for online vs. offline}

We notice that online distillation requires a smaller learning rate for stable training. Across all domains, LR=1e-6 yields better performance than LR=2e-5 (see Table \ref{tab:online_offline_optimal_lr}). In contrast, offline distillation exhibits the opposite trend, benefiting from a higher learning rate. This suggests that online training is more sensitive to learning rate selection, requiring careful tuning to avoid instability.

\begin{table}[h]
    \centering
    \begin{tabular}{l l c c}
        \toprule
        \textbf{Domain} & \textbf{Method} & \textbf{LR=1e-6} & \textbf{LR=2e-5} \\
        \midrule
        \multirow{2}{*}{Function Calling} & Online KL & {\bf 68.7 (+13.9\%)} & 65.4 (+8.5\%) \\
                      & Offline KL & 76.4 (+26.6\%) & {\bf 82.6 (+37.0\%)} \\
        \midrule
        \multirow{2}{*}{Biology}      & Online KL & {\bf 35.4 (+5.4\%)} & 33.6 (+0.1\%) \\
                     & Offline KL & 37.9 (+12.6\%) & {\bf 39.4 (+17.2\%)} \\
        \midrule
        \multirow{2}{*}{Chinese}      & Online KL & {\bf 30.2 (+2.5\%)} & 27.1 (-8.0\%) \\
                     & Offline KL & 31.0 (+5.0\%) & {\bf 33.6 (+14.0\%)} \\
        \bottomrule
    \end{tabular}
    \caption{Average acceptance rate (\%) and relative change from baseline for online and offline distillation across different domains.}
    \label{tab:online_offline_optimal_lr}
\end{table}

\subsection{Comparison of Forward KL and Reverse KL}
When evaluating different divergence measures, we notice that forward KL and reverse KL perform comparably at lower learning rates (1e-6). However, at a higher learning rate (2e-5) which yields better results for both losses, forward KL consistently outperforms reverse KL across different training sample sizes and domains (as shown in Figure \ref{fig:main results} and Figure \ref{fig:data_scaling}), making it the preferred choice for offline distillation.

\end{document}